\documentclass{article}
\usepackage{ijcai22}

\usepackage{times}
\usepackage{soul}
\usepackage{url}
\usepackage[hidelinks]{hyperref}
\usepackage[utf8]{inputenc}
\usepackage[small]{caption}
\usepackage{graphicx}
\usepackage{amsmath}
\usepackage{amssymb}
\usepackage{amsthm}
\usepackage{booktabs}
\usepackage{algorithm}
\usepackage{algorithmic}
\usepackage{multirow}
\usepackage{color}

\title{Distilling Inter-Class Distance for Semantic Segmentation}

\author{
Zhengbo Zhang$^1$
\and
Chunluan Zhou$^2$\and
Zhigang Tu$^{1}$\footnote{Corresponding author: Zhigang Tu}
\affiliations
$^1$Wuhan University\\
$^2$Wormpex AI Research
\emails
\{zhangzb, tuzhigang\}@whu.edu.cn,
czhou002@e.ntu.edu.sg
}

\begin{document}

\maketitle

\begin{abstract}
Knowledge distillation is widely adopted in semantic segmentation to reduce the computation cost. The previous knowledge distillation methods for semantic segmentation focus on pixel-wise feature alignment and intra-class feature variation distillation, neglecting to transfer the knowledge of the inter-class distance in the feature space, which is important for semantic segmentation. To address this issue, we propose an Inter-class Distance Distillation (IDD) method to transfer the inter-class distance in the feature space from the teacher network to the student network. Furthermore, semantic segmentation is a position-dependent task, thus we exploit a position information distillation module to help the student network encode more position information. Extensive experiments on three popular datasets: Cityscapes, Pascal VOC and ADE20K show that our method is helpful to improve the accuracy of semantic segmentation models and achieves the state-of-the-art performance. E.g. it boosts the benchmark model (``PSPNet+ResNet18") by 7.50$\%$ in accuracy on the Cityscapes dataset.
\end{abstract}

\section{Introduction}
Semantic segmentation aims at allocating a label for each pixel of the input image. It is a basic and challenging task in computer vision, which has widely applied in many fields, e.g. autonomous driving ~\cite{dong2020real}, ground feature changing detection~\cite{kemker2018algorithms}, etc. Recently, due to the success of deep learning ~\cite{tu2019action} in computer vision, Convolutional Neural Networks (CNNs) based methods have greatly improved the accuracy of semantic segmentation. However, CNN based semantic segmentation algorithms usually have an expensive computational cost, which limits their application in practice, especially for the real-life tasks that demand high efficiency.

To address this issue, many lightweight models have been explored, e.g. ENet ~\cite{paszke2016enet}, ESPNet~\cite{mehta2018espnet}, ICNet~\cite{zhao2018icnet}, and STDC~\cite{fan2021rethinking}. Although researchers have designed excellent networks to reduce the cost of computation, it is difficult to reach a satisfactory compromise between accuracy and model size. Instead of redesigning the backbone, we adopt the knowledge distillation (KD) strategy to train a student network by the guidance of a teacher network, and obtain comparable result.

\begin{figure}[!tp]
\includegraphics[width=1\linewidth]{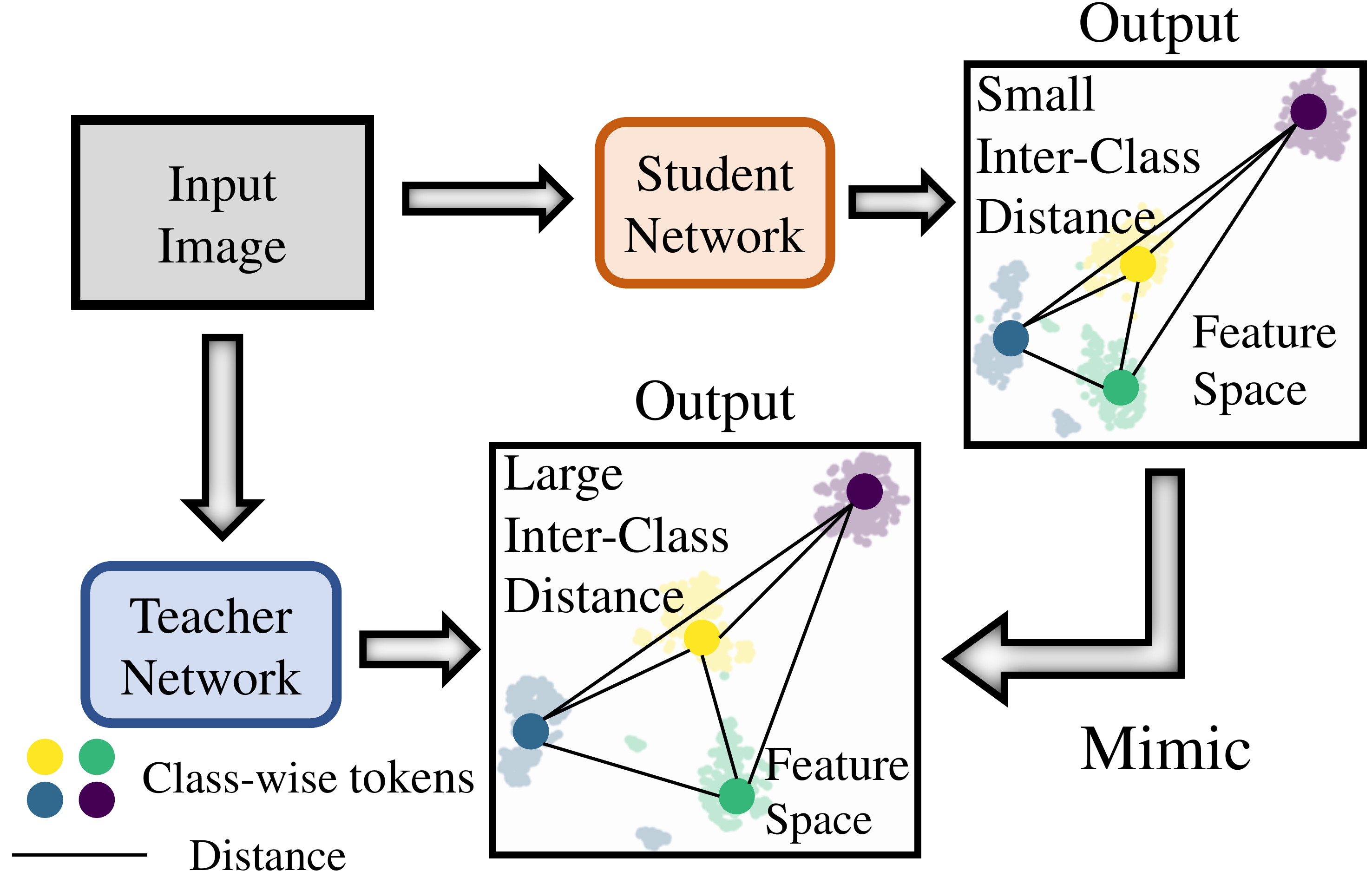}
  \centering
\caption{Limited by simple network structure and few parameters, the student network cannot have large inter-class distance like the teacher network in the feature space. Our motivation is to transfer the inter-class distance of the teacher network to help student network improve the segmentation accuracy.} 
\label{motivation}
\end{figure}

KD~\cite{hinton2015distilling}, as a model compression method, is originally used in the image classification task, which is able to simplify the cumbersome model significantly. Due to the advantage of KD, some semantic segmentation approaches use KD to reduce the model size~\cite{liu2019structured,wang2020intra,shu2021channel}. They force the student model to learn the pixel-wise feature and intra-class feature variation from the teacher network.  Representatively, Intra-Class Feature Variance Distillation (IFVD)~\cite{wang2020intra} focuses on transferring the variation of the intra-class feature from the teacher network to the student network. Channel-wise Knowledge Distillation (CD)~\cite{shu2021channel} emphasizes on distilling the most significant areas in each channel. 
It is worth noting that semantic segmentation is a pixel-wise category prediction task with various categories, thus the inter-class distance in the feature space is ubiquitous in semantic segmentation. Due to the help of numerous parameters and complex network structure, the teacher network has stronger classification ability and large inter-class distance in the feature space. However, \textit{Issue 1}: the past KD schemes for semantic segmentation neglect to transfer the inter-class distance in the feature space of the teacher network to the student network.   

Moreover, CNNs are able to encode the position information implicitly~\cite{islam2020much}. Semantic segmentation is a position-dependent task. Generally, with the simple network structure and a few parameters, \textit{Issue 2}: the student network is unable to encode as rich position information as the teacher network. 

To address the above mentioned issues, we consider to distill the inter-class distance in the feature space and position information from the teacher network to the student network. Accordingly, we propose a novel method (see Figure \ref{motivation}) called \emph{Inter-class Distance Distillation} (IDD). It consists of two main components. One is the \emph{inter-class distance distillation module} (IDDM), we design a graph to encode the inter-class distance, and make the student network mimic the large inter-class distance of the teacher network. The other is the \emph{position information distillation module} (PIDM). We design a position information network to extract the position information implicitly encoded in the feature map. Both the teacher network and the student network will predict the absolute coordinate mask via this network. By minimizing the divergence of them, the student network can encode more position information. With our IDD method, the student network learns more knowledge about inter-class distance and position information, improving the segmentation accuracy of the student network significantly.

The contributions are summarized in three-fold:
\begin{itemize}
\item We propose a novel approach named \emph{Inter-class Distance Distillation} (IDD) for semantic segmentation. It is the first method to distill the inter-class distance among all KD schemes for semantic segmentation to the best of our knowledge.

\item We design a \emph{position information distillation module} (PIDM) to enhance the capability of the student network encoding position information.

\item We demonstrate the effectiveness of the IDD method on three famous benchmark datasets, which not only obtains the state-of-the-art accuracy among KD schemes for semantic segmentation, but also is useful for other semantic segmentation models.
\end{itemize}

\section{Related Work}
\paragraph{Semantic segmentation.} 
CNN based models have greatly promoted the progress of semantic segmentation. Many researchers have tried different methods to make the model to learn rich contextual information. \cite{zhao2017pyramid} proposed a pyramid pooling strategy to collect context information from multiple scales. DeepLabv2~\cite{chen2017deeplab} adopted the atrous spatial pyramid pooling approach to get abundant context information. An encoder-decoder module was designed to capture multilevel features and contextual information. OCNet~\cite{yuan2018ocnet} exploited a self-attention mechanism to capture relationships between all pixels.
To meet the real-time semantic segmentation requirement of the mobile platform, some lightweight networks were proposed. ENet~\cite{paszke2016enet} used an asymmetric encoder-decoder structure and a convolution kernel decomposition operation, which greatly reduce the number of parameters and the floating point operations. Point-wise convolutions and spatial pyramid of dilated convolutions were applied in ESPNet~\cite{mehta2018espnet} to decrease the cost of computation. ICNet~\cite{zhao2018icnet} achieved fast semantic segmentation by designing an efficient network structure to process images with different resolutions. \cite{fan2021rethinking} designed a new real-time segmentation architecture by reducing network redundancy. Different from~\cite{mehta2018espnet,zhao2018icnet}, we get the lightweight semantic segmentation network with the usage of KD, which avoids to redesign the network structure, and gains high efficiency.

\paragraph{KD for semantic segmentation.} \cite{hinton2015distilling} proposed the concept of KD, it is a process of transferring the soft-labels from the teacher network to the student network to improve the performance of the student network. Because of the remarkable performance of KD, some researchers applied KD to semantic segmentation. ~\cite{liu2019structured} used a structured KD approach to transfer pixel-wise, pair-wise, and holistic knowledge from the teacher network. \cite{he2019knowledge} designed an autoencoder to transform knowledge into a compact form which is easier for the student network to learn. \cite{wang2020intra} presented an intra-class feature variation distillation scheme to make the student network simulate the intra-class feature distribution of the teacher network. \cite{shu2021channel} exploited a simple yet effective approach to minimize the channel-wise discrepancy between the teacher network and the student network. Unlike these mentioned approaches, our method pays attention to distilling the inter-class distance in the feature space, which is complementary to the previous distillation of pixel-wise feature alignment and intra-class feature variation.

\section{Proposed Method}
In this section, we first give an overview the general framework of past KD methods for semantic segmentation and our IDD model, then we describe the IDDM and PIDM in detail.

\subsection{Overview}
Semantic segmentation is a dense prediction task, aiming to assign a label to each pixel. Though the previous KD based semantic segmentation methods have achieved good progress, they mainly focus on aligning the pixel-wise feature and intra-class feature variance. Their loss function can be generally formulated as:
\begin{equation}
\begin{split}
\resizebox{.91\linewidth}{!}{$
\displaystyle
Loss = {L_{tar}}(D({\mathbf{GT}}),D(\mathbf{F}^S)) + \lambda  \cdot {L_{dis}}(\varphi(\mathbf{F}^T),\varphi(\mathbf{F}^S)),
$}\\
\resizebox{.91\linewidth}{!}{$
    \displaystyle
{L_{tar}}(D({\mathbf{GT}}),D(\mathbf{F}^S)) =  - \sum\limits_{k = 1}^N {D(\mathbf{GT}_k)\cdot\log (D(\mathbf{F}_k^S)).} 
$}
\end{split}
\end{equation}

\begin{figure*}[ht]
\includegraphics[width=0.7\linewidth]{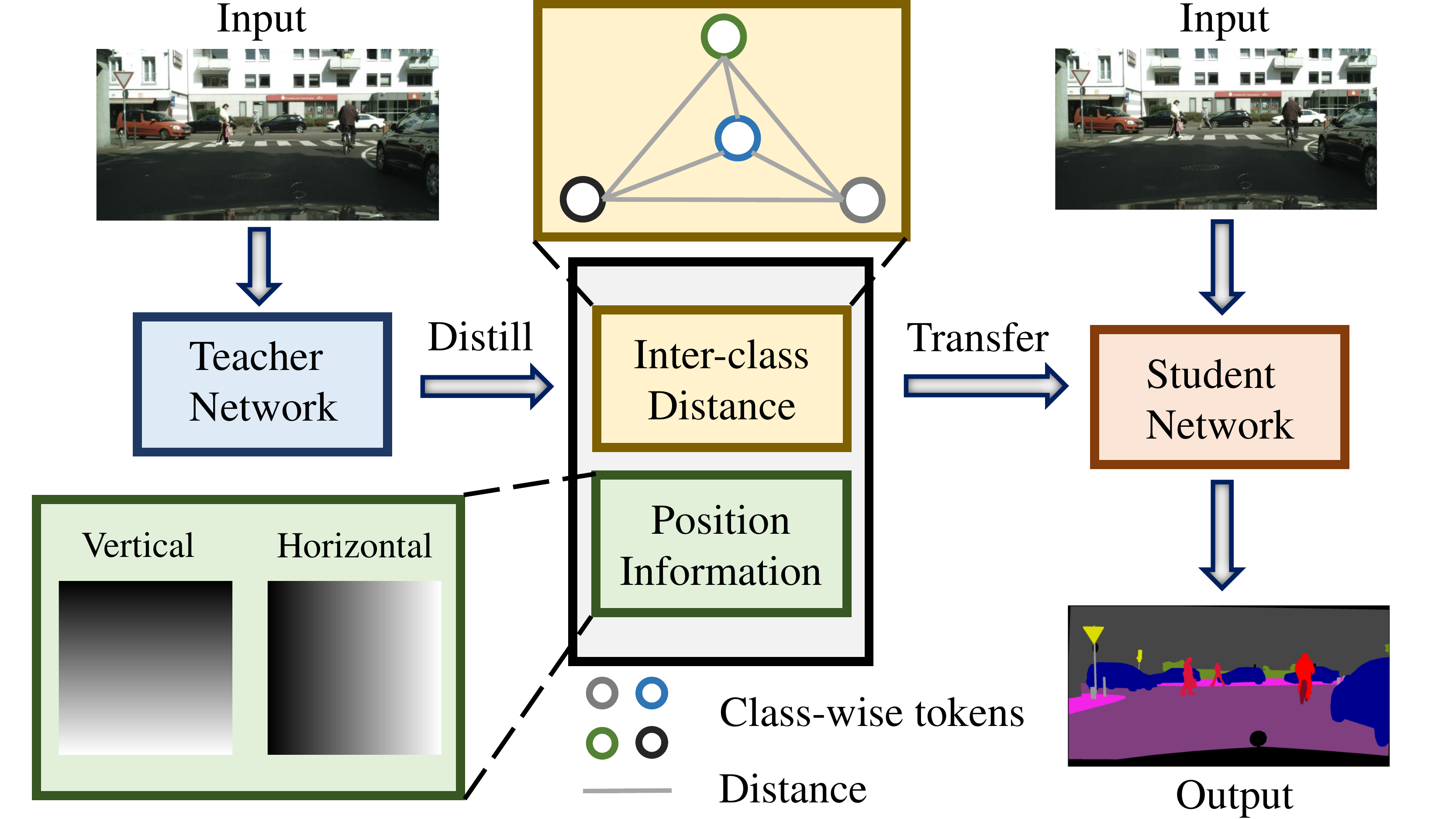}
  \centering
\caption{The network of our IDD method for semantic segmentation. We design a graph to encode the inter-class distance in the teacher network and transfer the inter-class distance to the student network. Besides, we transfer rich position information which is implicitly encoded in the teacher network to the student network.}
\label{network}
\end{figure*}

\noindent
${L_{tar}}$ is the cross-entropy loss, $\mathbf{GT}$ is the ground-truth, $\mathbf{F}^S$ and $\mathbf{F}^T$ denote the feature map of the student network and the teacher network, respectively. $\varphi( \cdot )$ represents a mapping function. $D({\mathbf{GT}})$ and $D({\mathbf{F}^S})$ separately denote the ground-truth and the student network's category probability distributions of all pixels. $N$ is the number of pixels, $D(\mathbf{GT}_k)$ denotes the $k^{th}$ pixel's ground-truth category probability distribution, $D(\mathbf{F}_k^{S})$ is the $k^{th}$ pixel's category probability distribution produced by the student network. $\lambda$ is a hyper-parameter to control the weight of loss. ${L_{dis}}( \cdot )$ is a loss function, such as the mean-squared error loss. Obviously, the prior methods ignore to transfer inter-class distance in the teacher network to the student network. Therefore, as illustrated in Figure \ref{network}, we propose the IDD method to transfer the inter-class distance and position information from the teacher to the student. We detail each module in the following subsections. 

\subsection{Inter-Class Distance Distillation Module}
Semantic segmentation is a pixel-wise classification task. Limited by simple network structure and few parameters, the student network has relatively poor discriminating ability and small inter-class distance. We propose the inter-class distance distillation module to deal with this challenge. 

As illustrated in Figure \ref{network}, we construct a graph $\mathcal{G}=\{\mathcal{V}, \mathcal{E}\}$ to encode the inter-class category distance, where $\mathcal{V} = \left \{ v_{i}\mid i=1,...,N   \right \} $ is a group of nodes, $N$ denotes the total number of segmentation categories of the processed image and $\mathcal{E} = \left \{ e_{i,j}\mid i=1,...N;j=1,...N;i \not=j   \right \}  $ represents a group of edges. $v_{i}$ denotes the token of the $i^{th}$ class, $v_{i}$ is obtained by averaging the feature of all pixels with the same category label $i$. $e_{i,j}$ is the Euclidean distance between the class-wise tokens of the $i^{th}$ and the $j^{th}$ category, which is defined as: %
\begin{align}
e_{i,j} = Dis(v_{i},v_{j}).
\end{align}
It represents the feature distance between the $i^{th}$ class and the $j^{th}$ class, and $Dis$ is the Euclidean distance. Due to the deep network and numerous parameters, the teacher network have large inter-class distance. Inspired by this characteristic, to enable the student network to better simulate the teacher network in terms of the inter-class distance, we design an inter-class distance loss function $L_{id}$, which is defined as:
\begin{align}
L_{id}=\frac{1}{2}\sum_{i=1}^{N}\sum_{j=1}^{N}  \left ( e_{i,j}^{T}-e_{i,j}^{S}  \right ) ^{2}, i \not=j,
\end{align}
where $e_{i,j}^{T}$ and $e_{i,j}^{S} $ stand for the $e_{i,j}$ in the teacher network and the student network, respectively.

\subsection{Position Information Distillation Module}

Semantic segmentation is a position-dependent task. It is reported in~\cite{islam2020much} that CNNs have the ability to encode position information. Inspired by~\cite{islam2020much}, we further introduce a position information distillation module to enhance the capability of the student network predicting position information. As a result, the student network can encode more position information in its output features which could be utilized to improve segmentation accuracy.

\begin{table}[ht]
\begin{center}
\begin{tabular}{cccccr}
\toprule
Backbone & ${L_{skd}}$ & ${L_{cw}}$ & ${L_{id}}$ & ${L_{pi}}$ & mIoU ($\%$) \\
\midrule
T: ResNet101 & & & & &78.56\\
S: ResNet18 & & & & &70.09\\
S: ResNet18 &\checkmark & & & &73.03\\
S: ResNet18 &\checkmark &\checkmark & & &75.78\\
S: ResNet18 &\checkmark &\checkmark &\checkmark & &76.81\\
S: ResNet18 &\checkmark &\checkmark & &\checkmark &76.43\\
S: ResNet18 &\checkmark &\checkmark &\checkmark &\checkmark &77.59\\
\bottomrule
\end{tabular}
\end{center}
\caption{Ablative studies of our loss items: ${L_{skd}}$, ${L_{cw}}$, ${L_{id}}$, and ${L_{pi}}$ on the Cityscapes validation dataset. ``T: ResNet101'' and ``S: ResNet18'' in the column of ``Backbone'' mean that we select ResNet101 and ResNet18 with PSPNet as the backbone for the teacher network and the student network, respectively.} 
\label{ablative study}
\end{table}

\begin{table}[ht]
\begin{tabular}{llllrrllrll}
\toprule
\multicolumn{4}{l}{{Method}} & {mIoU (\%)} & \multicolumn{3}{l}{{Params (M)}} & \multicolumn{3}{l}{{FLOPs (G)}} \\

\midrule
\multicolumn{4}{l}{ENet}                    & 58.3                      & \multicolumn{3}{r}{0.358}                      & \multicolumn{3}{r}{3.612}                     \\
\multicolumn{4}{l}{ESPNet}                  & 60.3                      & \multicolumn{3}{r}{0.364}                     & \multicolumn{3}{r}{4.422}                     \\
\multicolumn{4}{l}{ERFNet}                  & 68.0                      & \multicolumn{3}{r}{2.067}                      & \multicolumn{3}{r}{25.60}                     \\
\multicolumn{4}{l}{ICNet}                   & 69.5                      & \multicolumn{3}{r}{26.50}                      & \multicolumn{3}{r}{28.30}                     \\
\multicolumn{4}{l}{FCN}                     & 62.7                      & \multicolumn{3}{r}{134.5}                      & \multicolumn{3}{r}{333.9}                     \\
\multicolumn{4}{l}{RefineNet}               & 73.6                      & \multicolumn{3}{r}{118.1}                      & \multicolumn{3}{r}{525.7}                     \\
\multicolumn{4}{l}{OCNet}                   & 80.1                      & \multicolumn{3}{r}{62.58}                      & \multicolumn{3}{r}{548.5}                     \\ 
\multicolumn{4}{l}{T: PSPNet-R101}  & 78.4             & \multicolumn{3}{r}{70.43}             & \multicolumn{3}{r}{574.9}            \\ \midrule
\multicolumn{4}{l}{S: PSPNet-R18}   & 67.60          & \multicolumn{3}{r}{13.07}             & \multicolumn{3}{r}{125.8}            \\ 
\multicolumn{4}{l}{\textbf{S: +Ours (IDD)}}   & \textbf{76.33}            & \multicolumn{3}{r}{\textbf{13.07}}             & \multicolumn{3}{r}{\textbf{125.8}}            \\\bottomrule
\end{tabular}
\caption{Comparison the performance of different lightweight semantic segmentation models on the Cityscapes testing set.}
\label{methods comparison}
\end{table}

Specifically, we use $\mathbf{A}\in{\mathbb{R}^{C \times H \times W}}$ to represent the input feature map. First, we input $\mathbf{A}$ into a pretrained position information network to get the position information masks $\mathbf{P}^{HOR}\in{\mathbb{R}^{H \times W}}$ and $\mathbf{P}^{VER}\in{\mathbb{R}^{H \times W}}$, which represent the abscissa and ordinate respectively. In $\mathbf{P}^{HOR}$, each column has the same value, and we use $\mathbf{V}_j^{HOR}(j\in [1,H])$ to represent the value of column $j$, where $\mathbf{V}_j^{HOR} = j$. In $\mathbf{P}^{VER}$, each row has the same value, and we use $\mathbf{V}_i^{VER}(i\in [1,W])$ to denote the value of row $i$, where $\mathbf{V}_i^{VER} = i$.

We construct a loss function ${L_{pi}}$ to transfer the position information of the teacher network to the student network, it is expressed as:
\begin{align}
{L_{pi}} = \frac{1}{2} \cdot L_{pi}^{HOR} + \frac{1}{2} \cdot L_{pi}^{VER},
\end{align}
where
\begin{align}
\begin{split}
L_{pi}^{HOR} = \sum\limits_{j = 1}^H {{{\left\| {\frac{{Q_j^{HOR\_T}}}{{{{\left\| {Q_j^{HOR\_T}} \right\|}_2}}} - \frac{{Q_j^{HOR\_S}}}{{{{\left\| {Q_j^{HOR\_S}} \right\|}_2}}}} \right\|}_2,}}\\
L_{pi}^{VER} = \sum\limits_{i = 1}^W {{{\left\| {\frac{{Q_i^{VER\_T}}}{{{{\left\| {Q_i^{VER\_T}} \right\|}_2}}} - \frac{{Q_i^{HOR\_S}}}{{{{\left\| {Q_i^{HOR\_S}} \right\|}_2}}}} \right\|}_2}} 
\end{split}
\end{align}
\noindent
represent ${L_{pi}}$ in the horizontal and the vertical directions, respectively. ${Q_j^{VER\_T}}$ and ${Q_j^{VER\_S}}$ denote the column $j$ of $\mathbf{P}^{VER}$ produced by the teacher network and the student network in the vectorized form. Analogically, ${Q_i^{HOR\_T}}$ and ${Q_i^{HOR\_S}}$ denote the row $i$ of $\mathbf{P}^{HOR}$ produced by the teacher network and the student network in the vectorized form.

\subsection{Loss Function}
Following~\cite{shu2021channel}, we also apply the channel-wise supervision 
${L_{cw}}$ to minimize the Kullback–Leibler (KL) divergence of the channel-wise probability map between the teacher network and the student network. The final loss function of our IDD method is formulated as:
\begin{align}
L = {L_{skd}} + {\lambda _1} \cdot {L_{cw}} + {\lambda _2} \cdot {L_{id}} + {\lambda _3} \cdot {L_{pi}},
\end{align}
where ${L_{skd}}$ is a structured KD loss for semantic segmentation~\cite{liu2019structured}, ${\lambda _1}$, ${\lambda _2}$ and ${\lambda _3}$ are the hyper parameters to balance the weight between different items.

\section{Experiments}
To verify the effectiveness of our proposed IDD based semantic segmentation method, we conduct comprehensive experiments on three popular benchmarks:
Cityscapes~\cite{cordts2016cityscapes}, Pascal VOC~\cite{everingham2015pascal}, and ADE20K~\cite{zhou2017scene}. In the next subsections, we first introduce the datasets, evaluation metrics and implementation details. Next, we perform ablation experiments on the Cityscapes dataset. Finally, we compare our model with the state-of-the-art lightweight models on Cityscapes, Pascal VOC, and ADE20K.

\subsection{Datasets and Evaluation Metrics}

\paragraph{Datasets.} \textit{Cityscapes} includes 5000 finely annotated images of driving scenes in cities. It consists of 2975, 500 and 1525 images for training, validation and testing, respectively. It is labeled with 19 semantic categories. The resolution of each image is $2048 \times 1024$. In our experiments, we do not use the coarsely labeled images. \textit{Pascal VOC} composes of 1464 images for training, 1449 images for validation and 1456 images for testing. It covers 20 foreground object classes and 1 background class. \textit{ADE20k} is a challenging scene parsing dataset released by MIT, which contains 20K, 2K, 3K images with 150 classes for training, validation, and testing.

\paragraph{Evaluation metrics.} 
We use the Intersection-over-Union (IoU) of each class and the mean IoU (mIoU) of all classes to measure the segmentation accuracy.The total number of model parameters (Params) is utilized to measure the model size. We adopt an input image with resolution $512 \times 1024$ to calculate the floating-point operations per second (FLOPs), which is a general metric to measure the model complexity.

\subsection{Implementation Details}
\paragraph{Networks.} To make a fair and comparable evaluation, we carry out experiments on the same teacher and student network as~\cite{liu2019structured}. Specifically, in all of our experiments, PSPNet with ResNet101~\cite{he2016deep}, which are pretrained on ImageNet, is used as the teacher network. For the student network, we perform experiments on different segmentation architectures, such as the representative models PSPNet and Deeplab with the backbones of ResNet18 as well as ESPNet to verify the effectiveness of our IDD method. 

\begin{figure*}[ht]
\includegraphics[width=0.95\linewidth]{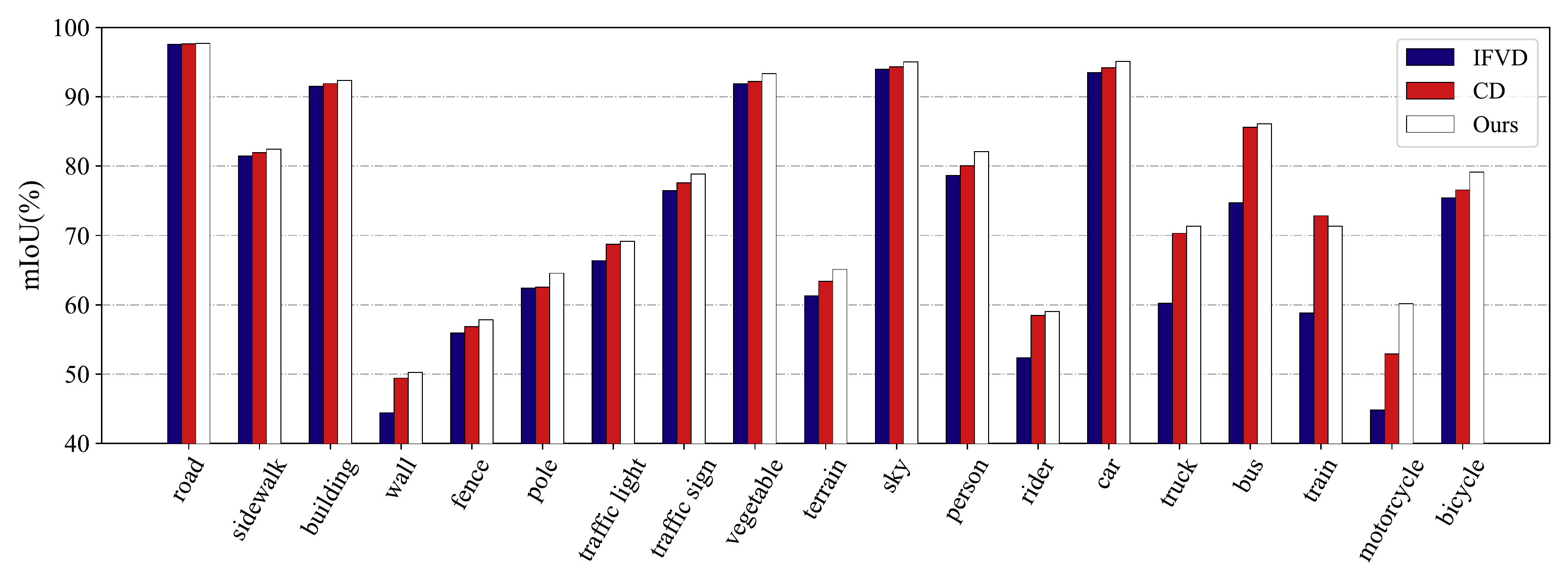}
  \centering
\caption{The class IoU scores of KD based semantic segmentation approaches on the Cityscapes validation dataset. We use PSPNet-R18(1.0) as the backbone of the student network.}
\label{classiou}
\end{figure*}

\begin{table}[ht]
\resizebox{245pt}{!}{
\begin{tabular}{llllrlrllrllrll}
\toprule
\multicolumn{4}{l}{\multirow{2}{*}{Method}} & \multicolumn{5}{r}{mIoU (\%)}                                             & \multicolumn{3}{r}{\multirow{2}{*}{Params (M)}} & \multicolumn{3}{r}{\multirow{2}{*}{FLOPs (G)}} \\ \cmidrule{5-9}
\multicolumn{4}{l}{}     & \multicolumn{2}{r}{Val}            & \multicolumn{3}{r}{Test}           & \multicolumn{3}{r}{}        & \multicolumn{3}{r}{}        \\ 
\midrule
\multicolumn{4}{l}{T: PSPNet-R101}           & \multicolumn{2}{r}{78.50}          & \multicolumn{3}{r}{78.40}          & \multicolumn{3}{r}{70.43}                      & \multicolumn{3}{r}{574.9}                     \\ \midrule
\multicolumn{4}{l}{S: ESPNet}                & \multicolumn{2}{r}{61.40}          & \multicolumn{3}{r}{60.30}          & \multicolumn{3}{r}{0.3635}                     & \multicolumn{3}{r}{4.422}                     \\
\multicolumn{4}{l}{+ SKD}                   & \multicolumn{2}{r}{63.80}          & \multicolumn{3}{r}{62.00}          & \multicolumn{3}{r}{0.3635}                     & \multicolumn{3}{r}{4.422}                     \\
\multicolumn{4}{l}{+ IFVD}                  & \multicolumn{2}{r}{65.13}          & \multicolumn{3}{r}{63.07}          & \multicolumn{3}{r}{0.3635}                     & \multicolumn{3}{r}{4.422}                     \\
\multicolumn{4}{l}{+ CD}                    & \multicolumn{2}{r}{67.27}          & \multicolumn{3}{r}{65.32}          & \multicolumn{3}{r}{0.3635}                     & \multicolumn{3}{r}{4.422}                     \\
\multicolumn{4}{l}{\textbf{+ Ours}}         & \multicolumn{2}{r}{\textbf{68.87}} & \multicolumn{3}{r}{\textbf{67.35}} & \multicolumn{3}{r}{\textbf{0.3635}}            & \multicolumn{3}{r}{\textbf{4.422}}            \\ \midrule
\multicolumn{4}{l}{S: PSPNet-R18 (0.5)}       & \multicolumn{2}{r}{61.17}          & \multicolumn{3}{r}{-}          & \multicolumn{3}{r}{3.271}                      & \multicolumn{3}{r}{31.53}                     \\
\multicolumn{4}{l}{+ SKD}                   & \multicolumn{2}{r}{61.60}          & \multicolumn{3}{r}{60.05}          & \multicolumn{3}{r}{3.271}                      & \multicolumn{3}{r}{31.53}                     \\
\multicolumn{4}{l}{+ IFVD}                  & \multicolumn{2}{r}{63.35}          & \multicolumn{3}{r}{63.68}          & \multicolumn{3}{r}{3.271}                      & \multicolumn{3}{r}{31.53}                     \\
\multicolumn{4}{l}{+ CD}                    & \multicolumn{2}{r}{68.57}          & \multicolumn{3}{r}{66.75}          & \multicolumn{3}{r}{3.271}                      & \multicolumn{3}{r}{31.53}                     \\
\multicolumn{4}{l}{\textbf{+ Ours}}         & \multicolumn{2}{r}{\textbf{69.76}} & \multicolumn{3}{r}{\textbf{68.54}} & \multicolumn{3}{r}{\textbf{3.271}}             & \multicolumn{3}{r}{\textbf{31.53}}            \\ \midrule
\multicolumn{4}{l}{S: PSPNet-R18}            & \multicolumn{2}{r}{70.09}          & \multicolumn{3}{r}{67.60}          & \multicolumn{3}{r}{13.07}                      & \multicolumn{3}{r}{125.8}                     \\
\multicolumn{4}{l}{+ SKD}                   & \multicolumn{2}{r}{72.70}          & \multicolumn{3}{r}{71.40}          & \multicolumn{3}{r}{13.07}                      & \multicolumn{3}{r}{125.8}                     \\
\multicolumn{4}{l}{+ IFVD}                  & \multicolumn{2}{r}{74.54}          & \multicolumn{3}{r}{72.74}          & \multicolumn{3}{r}{13.07}                      & \multicolumn{3}{r}{125.8}                     \\
\multicolumn{4}{l}{+ CD}                    & \multicolumn{2}{r}{75.90}          & \multicolumn{3}{r}{74.58}          & \multicolumn{3}{r}{13.07}                      & \multicolumn{3}{r}{125.8}                     \\
\multicolumn{4}{l}{\textbf{+ Ours}}         & \multicolumn{2}{r}{\textbf{77.59}} & \multicolumn{3}{r}{\textbf{76.33}} & \multicolumn{3}{r}{\textbf{13.07}}             & \multicolumn{3}{r}{\textbf{125.8}}            \\ \bottomrule

\end{tabular}
}
\caption{Comparison of different KD based semantic segmentation methods on the Cityscapes dataset. ``PSPNet-R18(0.5)" is trained from scratch.}
\label{cityscapes compare}
\end{table}

\paragraph{Training Details.} We use the Pytorch platform to implement our method. Following~\cite{liu2019structured} , we train our student networks by mini-batch stochastic gradient descent (SGD) for $40000$ iterations. We set the momentum and the weight decay as 0.9 and 0.0005, respectively. We apply the polynomial learning rate policy, and the learning rate is calculated as $base\_lr\cdot{\left( {1 - \frac{{iter}}{{total\_iter}}} \right)^{power}}$. The base learning rate and power are respectively set to 0.01 and 0.9. For the input images, we crop them to $512 \times 512$. The random scaling and random flipping are applied to augment the data.

\begin{figure}[ht]
\includegraphics[width=0.95\linewidth]{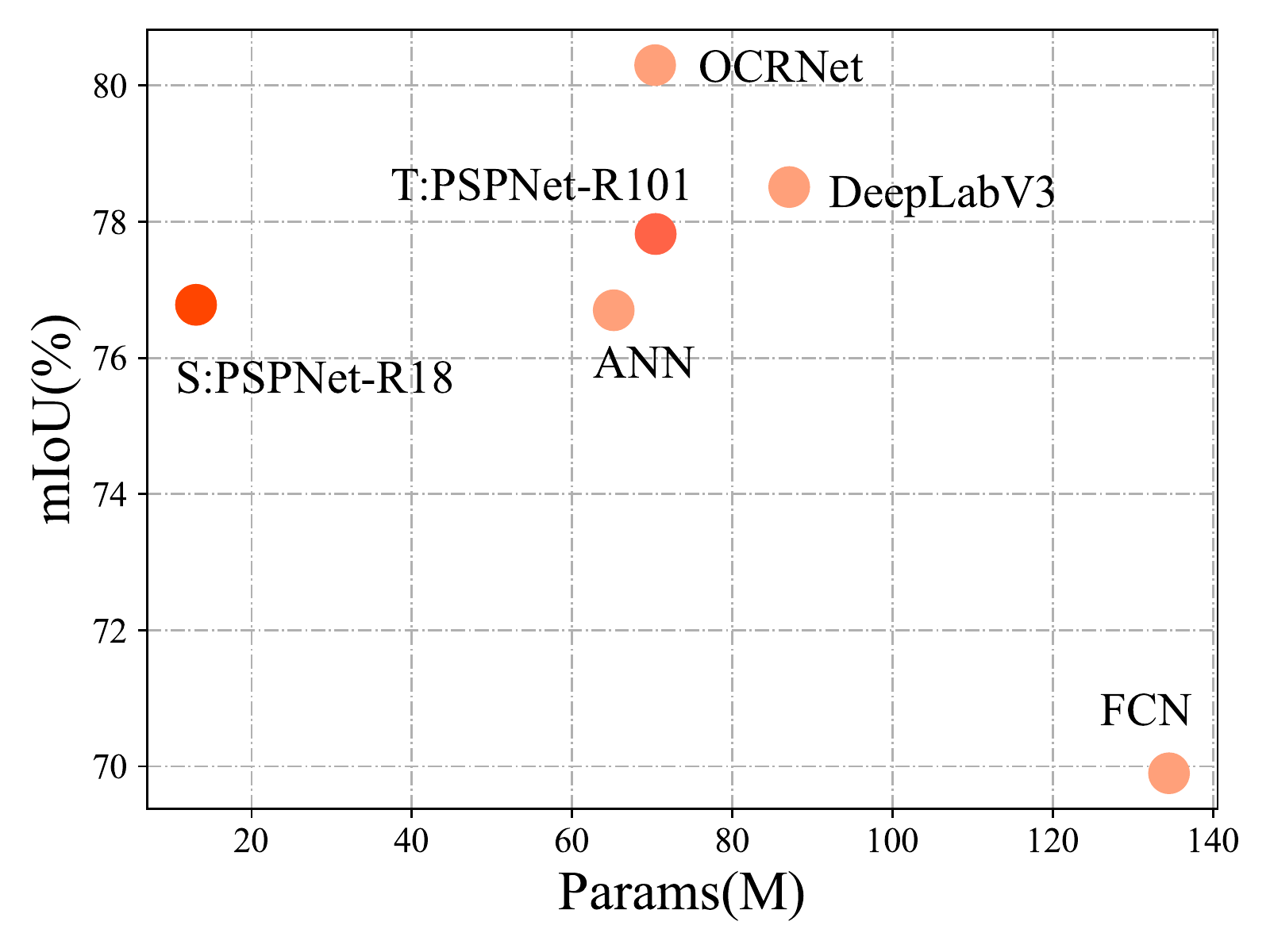}
  \centering
\caption{Comparison of Params and mIoU of different models on the Pascal VOC validation set. We use PSPNet-R18(1.0) as the backbone of the student network.}
\label{pascalvoc}
\end{figure}

\subsection{Ablative Study}
Our loss function consists of four parts, ${L_{skd}}$, ${L_{cw}}$, ${L_{id}}$, and ${L_{pi}}$. To explore the effectiveness of each loss item, we conduct ablation experiments on the Cityscapes validation dataset with the evaluation metric mIoU {(\%)}. The teacher network is PSPNet~\cite{zhao2017pyramid} with ResNet101 backbone  (``T: PSPNet-R101''), and the student model is PSPNet with ResNet18 (``S: PSPNet-R18'') also pretrained in the ImageNet. As can be seen in Table \ref{ablative study}, the structured KD loss ${L_{skd}}$ boosts the performance of the student network ``S: PSPNet-R18'' from 70.09{\%} to 73.03{\%}. The channel-wise KD loss ${L_{cw}}$ further improves the student model to 75.78{\%}. By adopting our inter-class distance distillation approach, the gain increases to 5.34{\%} (76.43\% vs 70.09{\%}). Furthermore, after applying our position information loss ${L_{pi}}$, the accuracy of the lightweight student network ``S: PSPNet-R18'' reaches 77.59{\%}, approximately to the accuracy of the teacher network ``T: PSPNet-R101'', the mIoU value of which is 78.56\%. The experimental results prove that our proposed IDDM and PIDM are effective.

\begin{figure}[ht]
\includegraphics[width=0.9\linewidth]{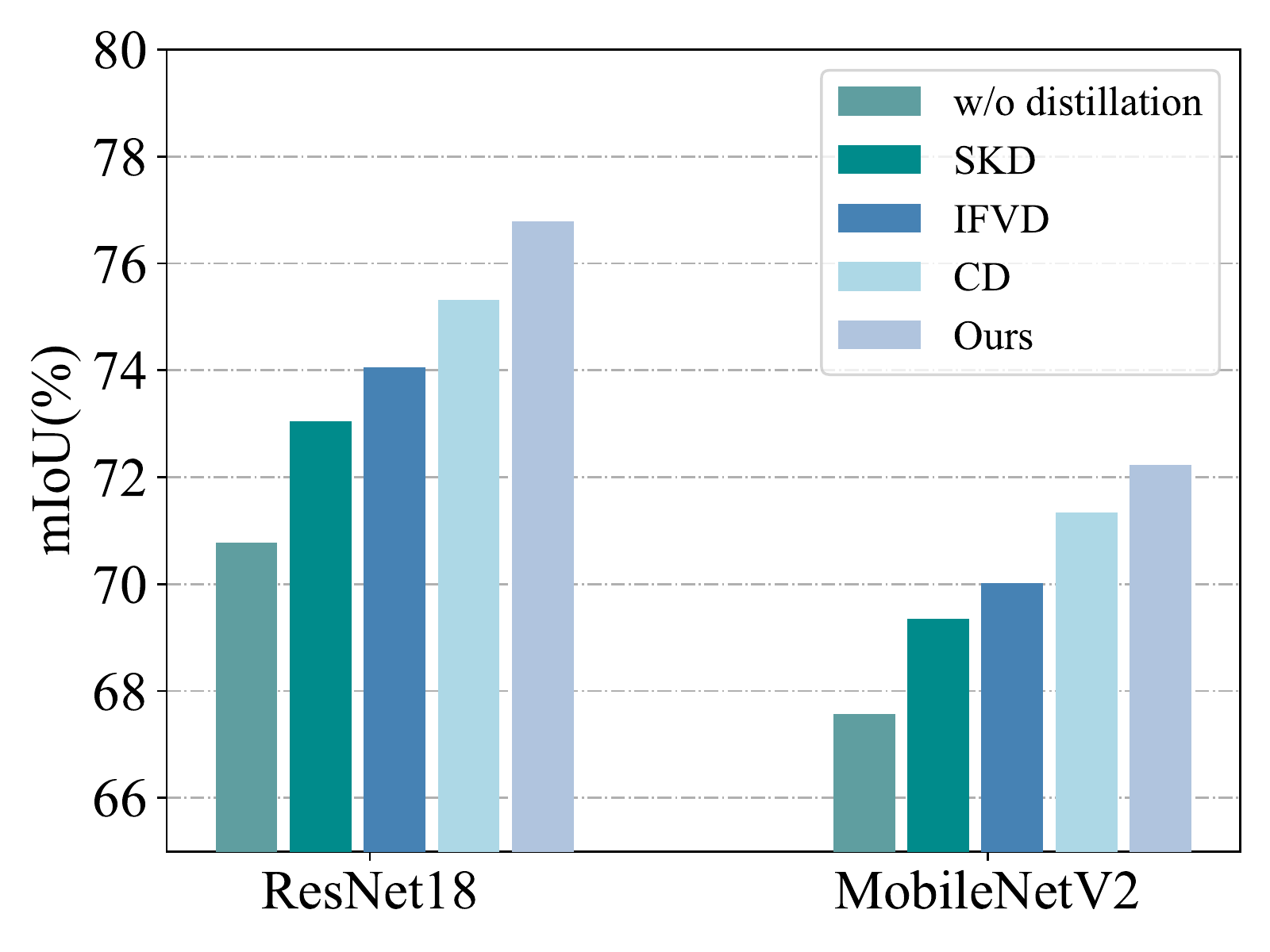}
  \centering
\caption{Comparison of different KD strategies for semantic segmentation on the Pascal VOC validation dataset.}
\label{pascalvoc_zhu}
\end{figure}

\begin{table}[th]
\centering
\begin{tabular}{llllrlllrll}
\toprule
\multicolumn{4}{l}{Method}        & \multicolumn{4}{r}{mIoU (\%)} & \multicolumn{3}{r}{Params (M)} \\\midrule
\multicolumn{4}{l}{T:PSPNet-R101} & \multicolumn{4}{r}{44.94}     & \multicolumn{3}{r}{70.43}      \\ \midrule
\multicolumn{4}{l}{S:PSPNet-R18}  & \multicolumn{4}{r}{24.65}     & \multicolumn{3}{r}{13.07}      \\
\multicolumn{4}{l}{+SKD}          & \multicolumn{4}{r}{25.02}     & \multicolumn{3}{r}{13.07}      \\
\multicolumn{4}{l}{+IFVD}         & \multicolumn{4}{r}{25.82}     & \multicolumn{3}{r}{13.07}      \\
\multicolumn{4}{l}{+CD}           & \multicolumn{4}{r}{26.80}     & \multicolumn{3}{r}{13.07}      \\
\multicolumn{4}{l}{\textbf{+Ours}}         & \multicolumn{4}{r}{\textbf{27.69}}     & \multicolumn{3}{r}{\textbf{13.07}}      \\ \midrule
\multicolumn{4}{l}{S:PSPNet-MNV2} & \multicolumn{4}{r}{23.21}     & \multicolumn{3}{r}{2.15}       \\
\multicolumn{4}{l}{+SKD}          & \multicolumn{4}{r}{24.89}     & \multicolumn{3}{r}{2.15}       \\
\multicolumn{4}{l}{+IFVD}         & \multicolumn{4}{r}{25.43}     & \multicolumn{3}{r}{2.15}       \\
\multicolumn{4}{l}{+CD}           & \multicolumn{4}{r}{27.74}     & \multicolumn{3}{r}{2.15}       \\
\multicolumn{4}{l}{\textbf{+Ours}}         & \multicolumn{4}{r}{\textbf{28.93}}     & \multicolumn{3}{r}{\textbf{2.15}}       \\ \bottomrule
\end{tabular}
\caption{Comparison of different KD approaches for semantic segmentation methods on the ADE20K validation dataset.}
\label{ADE20Kcompare}
\end{table}

\subsection{Results}
\subsubsection{Cityscapes}                            
Table \ref{methods comparison} shows the quantitative results on the Cityscapes dataset. By using our IDD, the Params and the FLOPs of the our student network (``\textbf{Ours}") reduce by 81.44$\%$ (13.07 vs 70.43) and 78.12$\%$ (125.8 vs 574.9) compared to the teacher network, while the mIoU accuracy only decreases 2.07\% (from 78.4$\%$ to 76.33$\%$). Compared with other lightweight models, our method also has remarkable performance. For example, our IDD outperforms ENet~\cite{paszke2016enet} and ESPNet~\cite{mehta2018espnet} by 18.03$\%$ and 16.03$\%$ in accuracy (mIoU), respectively. Notably, the Params of ours are only half of ICNet~\cite{zhao2018icnet}, but the accuracy of our student network is still 5.0$\%$ higher. Although the accuary of OCNet~\cite{yuan2018ocnet} is 3.77$\%$ higher than ours, the Params of ours are less than one fifth of OCNet. The results demonstrate that IDD achieves a satisfactory compromise between accuracy and model size.

We also evaluate the performance of our method and other KD based methods on the Cityscapes, e.g. SKD~\cite{he2019knowledge}, IFVD~\cite{wang2020intra} and CD~\cite{shu2021channel}. The  student models are ESPNet, PSPNet-R18(0.5) and PSPNet-R18. Experimental results are listed in Table \ref{cityscapes compare}. When we adopt ESPNet as the student network, our method leads to a significant improvement of 7.47$\%$ and 7.05$\%$ on the validation set and the testing set, respectively. Compared with SKD which transfers the intra-class feature variance and CD which transfers the channel-wise feature, our method outperforms them by 3.74$\%$ and 1.60$\%$, separately. 
After using our IDD, the performance of PSPNet-R18(0.5) increases from 61.17$\%$ to 69.76$\%$, and surpasses IFVD and CD by 6.41$\%$ and 1.19$\%$ in the validation set. When PSPNet-R18 is adopted as the student model, with our IDD, the gains reach to 7.50$\%$ (70.09$\%$ to 77.59$\%$), and outperform IFVD and CD by 3.05$\%$ and 1.69$\%$ respectively. The experimental results show that our IDD is better than the previous KD strategies for semantic segmentation. 

In addition, as shown in Figure \ref{classiou}, we use the PSPNet-R18(1.0) as the student network to calculate the mIoU for each class compared with two state-of-the-art methods. Due to our method enables the student network to have large inter-class distance and rich position information, it performs well on some categories. For example, rider, car and bus. Table \ref{cityscapes compare} shows the qualitative results, which again demonstrate the effectiveness of our IDD method.

\subsubsection{Pascal VOC}

As depicted in Figure \ref{pascalvoc}, we adopt a dot graph to describe the parameters and accuracy of different networks, i.e. OCRNet~\cite{yuan2020object}, DeepLabV3, FCN~\cite{long2015fully}, ANN~\cite{zhu2019asymmetric} and PSPNet. By using our spatial knowledge distillation, the PSPNet-R18(1.0) outperforms FCN and ANN by 6.79$\%$ and 0.08$\%$, respectively. 

We adopt ResNet18 and MobileNetV2 as the student network to evaluate our approach on the validation set. The results are shown in Figure~\ref{pascalvoc_zhu}. With ResNet18 as the backbone of the student network, our approach improves the accuracy of the model that without distillation by 6.01$\%$, and is better than the SKD, IFVD and CD respectively by 3.74$\%$, 2.74$\%$ and 1.47$\%$. For MobileNetV2, our method exceeds the benchmark model by 4.66$\%$, and improves the SKD, IFVD and CD respectively by 2.88$\%$, 2.21$\%$ and 0.89$\%$.

\subsubsection{ADE20K}
To further verify the effectiveness of our proposed method, we carry out experiments on the challenging dataset ADE20K. The quantitative results are reported in Table \ref{ADE20Kcompare}. When the student model is built on ResNet18, our proposed approach improves the student model from 24.65$\%$ to 27.65$\%$, and outperforms SKD, IFVD and CD by 2.67$\%$, 1.87$\%$ and 0.89$\%$. With MobileNetV2 as the student backbone, we achieve an improvement to 6.72$\%$ compared with the benchmark model, and improves the SKD, IFVD and CD by 4.04$\%$, 3.50$\%$ and 1.19$\%$, respectively.

\section{Conclusion}
In this paper, we present a novel knowledge distillation method for semantic segmentation, helping the student model have large inter-class distance in the feature space and rich position information. Specifically, we propose the inter-class distance distillation module and the position information distillation module to transfer the inter-class distance and position cue from the teacher network to the student network. Ablative experiments show that our explored two modules enable the student network to mimic the teacher network better. We demonstrate the effectiveness of our approach by conducting extensive experiments on three public datasets, i.e. Cityscapes, Pascal VOC and ADE20K.

\section*{Acknowledgements}
This work was supported by the National Natural Science Foundation of China under Grant 62106177. The numerical calculation was supported by the super-computing system in the Super-computing Center of Wuhan University.

{
\bibliographystyle{named}
\bibliography{ijcai22}
}

\end{document}